\documentclass{article}
\PassOptionsToPackage{numbers, compress}{natbib}
\usepackage[final]{neurips_2019}

\usepackage[utf8]{inputenc} 
\usepackage[T1]{fontenc}    
\usepackage{url}            
\usepackage{booktabs}       
\usepackage{amsfonts}       
\usepackage{amsmath}
\usepackage{nicefrac}       
\usepackage{microtype}      
\usepackage{algorithm}
\usepackage{algorithmic}
\usepackage[hidelinks]{hyperref}
\usepackage{graphicx}
\usepackage{subfigure}
\usepackage{placeins}
\usepackage{todonotes}
\usepackage[group-separator={,}]{siunitx}
\usepackage{xspace}

\graphicspath{{./figures/}}

\newcommand{\ALG}{\texttt{TuRBO}\xspace}

\newcommand{\HeSBO}{\texttt{HeSBO-TS}\xspace}
\newcommand{\BOCK}{\texttt{BOCK}\xspace}
\newcommand{\EBO}{\texttt{EBO}\xspace}
\newcommand{\NM}{\texttt{NM}\xspace}
\newcommand{\CMA}{\texttt{CMA-ES}\xspace}
\newcommand{\TS}{\texttt{TS}\xspace}
\newcommand{\GPTS}{\texttt{GP-TS}\xspace}
\newcommand{\COBYLA}{\texttt{COBYLA}\xspace}
\newcommand{\BOBYQA}{\texttt{BOBYQA}\xspace}
\newcommand{\RS}{\texttt{RS}\xspace}
\newcommand{\BFGS}{\texttt{BFGS}\xspace}
\newcommand{\OpenAI}{\texttt{OpenAI}\xspace}
\newcommand{\BOHAMIANN}{\texttt{BOHAMIANN}\xspace}
\newcommand{\narm}{\ensuremath{m}\xspace}

\renewcommand{\vec}[1]{{\boldsymbol{\mathbf{#1}}}} 

\DeclareMathOperator*{\argmin}{argmin}

\newcommand{\xbest}{\vec x^\star}

\newcommand{\len}{L}

\newcommand{\nbatch}{q}

\title{Scalable Global Optimization via\\Local Bayesian Optimization}
\author{
  David Eriksson \\
  Uber AI \\
  \texttt{eriksson@uber.com} \\
  \And
  Michael Pearce$^*$ \\
  University of Warwick \\
  \texttt{m.a.l.pearce@warwick.ac.uk} \\
  \And
  Jacob R Gardner \\
  Uber AI \\
  \texttt{jake.gardner@uber.com} \\
  \And
  Ryan Turner \\
  Uber AI \\
  \texttt{ryan.turner@uber.com} \\
  \And
  Matthias Poloczek \\
  Uber AI \\
  \texttt{poloczek@uber.com} \\
}

\begin{document}

{\let\thefootnote\relax\footnote{
  {$^*$This work was conducted while Michael Pearce was visiting Uber AI.}}}

\maketitle

\begin{abstract}
Bayesian optimization has recently emerged as a popular method for the sample-efficient optimization of expensive black-box functions.
However, the application to high-dimensional problems with several thousand observations remains challenging, and on difficult problems Bayesian optimization is often not competitive with other paradigms.
In this paper we take the view that this is due to the implicit homogeneity of the global probabilistic models and an overemphasized exploration that results from global acquisition.
This motivates the design of a \emph{local} probabilistic approach for global optimization of large-scale high-dimensional problems.
We propose the \ALG algorithm that fits a collection of local models and performs a principled global allocation of samples across these models via an implicit bandit approach.
A comprehensive evaluation demonstrates that \ALG outperforms state-of-the-art methods from machine learning and operations research on problems spanning reinforcement learning, robotics, and the natural sciences.

\end{abstract}

\section{Introduction}
\label{sec:introduction}
The global optimization of high-dimensional black-box functions---where closed form expressions and derivatives are unavailable---is a ubiquitous task arising in hyperparameter tuning \cite{snoek2012practical}; in reinforcement learning, when searching for an optimal parametrized policy \citep{calandra2016bayesian}; in simulation, when calibrating a simulator to real world data; and in chemical engineering and materials discovery, when selecting candidates for high-throughput screening \citep{hernandez2017parallel}.
While Bayesian optimization (BO) has emerged as a highly competitive tool for problems with a small number of tunable parameters (e.g., see \citep{frazier2018tutorial,shahriari2016taking}), it often scales poorly to high dimensions and large sample budgets.
Several methods have been proposed for high-dimensional problems with small budgets of a few hundred samples (see the literature review below). However, these methods make strong assumptions about the objective function such as low-dimensional subspace structure.
The recent algorithms of~\citet{wang2017batched} and~\citet{hernandez2017parallel} are explicitly designed for a large sample budget and do not make these assumptions.
However, they do not compare favorably with state-of-the-art methods from stochastic optimization like \CMA \citep{hansen2006cma} in practice.

The optimization of high-dimensional problems is hard for several reasons.
First, the search space grows exponentially with the dimension, and while local optima may become more plentiful, global optima become more difficult to find.
Second, the function is often heterogeneous, making the task of fitting a global surrogate model challenging.
For example, in reinforcement learning problems with sparse rewards, we expect the objective function to be nearly constant in large parts of the search space.
For the latter, note that the commonly used global Gaussian process (GP) models \citep{frazier2018tutorial,williams2006gaussian} implicitly suppose that characteristic lengthscales and signal variances of the function are constant in the search space.
Previous work on non-stationary kernels does not make this assumption, but these approaches are too computationally expensive to be applicable in our large-scale setting \cite{snoek2014input,taddy2009bayesian,assael2014heteroscedastic}.
Finally, the fact that search spaces grow considerably faster than sampling budgets due to the curse of dimensionality implies the inherent presence of regions with large posterior uncertainty.
For common myopic acquisition functions, this results in an overemphasized exploration and a failure to exploit promising areas.

To overcome these challenges, we adopt a \emph{local} strategy for BO.
We introduce trust region BO (\ALG), a technique for global optimization, that uses a collection of simultaneous \emph{local} optimization runs using independent probabilistic models.
Each local surrogate model enjoys the typical benefits of Bayesian modeling ---robustness to noisy observations and rigorous uncertainty estimates--- however, these local surrogates allow for heterogeneous modeling of the objective function and do not suffer from over-exploration.
To optimize globally, we leverage an implicit multi-armed bandit strategy at each iteration to allocate samples between these local areas and thus decide which local optimization runs to continue.

We provide a comprehensive experimental evaluation demonstrating that \ALG outperforms the state-of-the-art from BO, evolutionary methods, simulation optimization, and stochastic optimization on a variety of benchmarks that span from reinforcement learning to robotics and natural sciences.
An implementation of~\ALG is available at \url{https://github.com/uber-research/TuRBO}.

\subsection{Related work}
\label{sec:background}
BO has recently become the premier technique for global optimization of expensive functions, with applications in hyperparameter tuning, aerospace design, chemical engineering, and materials discovery; see \citep{frazier2018tutorial,shahriari2016taking} for an overview.
However, most of BO's successes have been on low-dimensional problems and small sample budgets.
%
This is not for a lack of trying; there have been many attempts to scale BO to more dimensions and observations.
A common approach is to replace the GP model: \citet{hutter2011sequential} uses random forests, whereas \citet{snoek2015scalable} applies Bayesian linear regression on features from neural networks.
This neural network approach was refined by \citet{springenberg2016bayesian} whose \BOHAMIANN algorithm uses a modified Hamiltonian Monte Carlo method, which is more robust and scalable than standard Bayesian neural networks.
\citet{hernandez2017parallel} combines Bayesian neural networks with Thompson sampling (\TS), which easily scales to large batch sizes. We will return to this acquisition function later.

There is a considerable body of work in high-dimensional BO~\citep{chen2012joint,kandasamy2015high,binois2015warped,wang2016bayesian,gardner2017discovering,wang2017batched,rolland2018high,mutny2018efficient,HeSBO19,binois2017choice}. Many methods exist that exploit potential additive structure in the objective function~\citep{kandasamy2015high,gardner2017discovering,wang2017batched}. These methods typically rely on training a large number of GPs (corresponding to different additive structures) and therefore do not scale to large evaluation budgets. Other methods exist that rely on a mapping between the high-dimensional space and an unknown low-dimensional subspace to scale to large numbers of observations \citep{wang2016bayesian,HeSBO19,garnett2013active}.
The \BOCK algorithm of~\citet{oh2018bock} uses a cylindrical transformation of the search space to achieve scalability to high dimensions.
\emph{Ensemble Bayesian optimization} (\EBO)~\citep{wang2017batched} uses an ensemble of additive GPs together with a batch acquisition function to scale BO to tens of thousands of observations and high-dimensional spaces.
Recently, \citet{HeSBO19} have proposed the general \texttt{HeSBO} framework that extends GP-based BO algorithms to high-dimensional problems using a novel subspace embedding that overcomes the limitations of the Gaussian projections used in \citep{wang2016bayesian,binois2015warped,binois2017choice}.
From this area of research, we compare to \BOCK, \BOHAMIANN, \EBO, and \texttt{HeSBO}.

To acquire large numbers of observations, large-scale BO usually selects points in batches to be evaluated in parallel.
While several batch acquisition functions have recently been proposed~\citep{chevalier2013fast,shah2015parallel,wang2016parallel,wu2016parallel,wu2017bayesian,marmin2015differentiating,gonzalez2016batch}, these approaches do not scale to large batch sizes in practice.
\TS~\citep{thompson1933likelihood} is particularly lightweight and easy to implement as a batch acquisition function as the computational cost scales linearly with the batch size.
Although originally developed for bandit problems~\citep{russo2018tutorial}, it has recently shown its value in BO~\citep{hernandez2017parallel,baptista2018bayesian,kandasamy2018parallelised}.
In practice, \TS is usually implemented by drawing a realization of the unknown objective function from the surrogate model's posterior on a discretized search space.
Then, \TS finds the optimum of the realization and evaluates the objective function at that location.
This technique is easily extended to batches by drawing multiple realizations as (see the supplementary material for details).

Evolutionary algorithms are a popular approach for optimizing black-box functions when thousands of evaluations are available, see \citet{jin2005evolutionary} for an overview in stochastic settings.
We compare to the successful covariance matrix adaptation evolution strategy (\CMA) of \citet{hansen2006cma}.
\CMA performs a stochastic search and maintains a multivariate normal sampling distribution over the search space.
The evolutionary techniques of recombination and mutation correspond to adaptions of the mean and covariance matrix of that distribution.

%
High-dimensional problems with large sample budgets have also been studied extensively in operations research and simulation optimization, see \citep{dong2017empirically} for a survey.
Here the successful trust region (TR) methods are based on a local surrogate model in a region (often a sphere) around the best solution.
The trust region is expanded or shrunk depending on the improvement in obtained solutions; see \citet{Yuan2000} for an overview.
We compare to \BOBYQA~\citep{powell2007view}, a state-of-the-art TR method that uses a quadratic approximation of the objective function.
We also include the Nelder-Mead (\NM) algorithm~\citep{nelder1965simplex}.
For a~$d$-dimensional space, \NM creates a~$(d+1)$-dimensional simplex that adaptively moves along the surface by projecting the vertex of the worst function value through the center of the simplex spanned by the remaining vertices.
Finally, we also consider the popular quasi-Newton method BFGS~\citep{Zhu1997}, where gradients are obtained using finite differences.
For other work that uses local surrogate models, see e.g., \cite{krityakierne2015global,wabersich2016advancing,acerbi2017practical,akrour2017local,mcleod2018optimization}.

\section{The trust region Bayesian optimization algorithm}
\label{sec:methods}
In this section, we propose an algorithm for optimizing high-dimensional black-box functions.
In particular, suppose that we wish to solve:
\begin{equation*}
    \text{Find } \vec x^* \in \Omega \text{ such that } f( \vec x^*) \leq f(\vec x),
    \,\,\,\forall \vec x \in \Omega,
\end{equation*}
where \mbox{$f : \Omega \to \mathbb{R}$} and $\Omega = [0, 1]^d$. We observe potentially noisy values
$y(\vec x) = f(\vec x) + \varepsilon$, where $\varepsilon \sim \mathcal{N}(0, \sigma^2)$.
BO relies on the ability to construct a global model that is \emph{eventually} accurate enough to uncover a global optimizer. As discussed previously, this is challenging due to the curse of dimensionality and the heterogeneity of the function.
To address these challenges, we propose to abandon global surrogate modeling,
and achieve global optimization by maintaining several \emph{independent local models}, each involved in a separate local optimization run.
To achieve global optimization in this framework, we maintain multiple local models simultaneously and allocate samples via an implicit multi-armed bandit approach.
This yields an efficient acquisition strategy that directs samples towards promising local optimization runs.
We begin by detailing a single local
optimization run, and then discuss how multiple runs are managed.

\paragraph{Local modeling.}
To achieve principled local optimization in the gradient-free setting, we draw inspiration from a class of TR methods from stochastic optimization \citep{Yuan2000}. These methods make suggestions using a (simple) surrogate model inside a TR. The region is often a sphere or a polytope centered at the best solution, within which the surrogate model is believed to accurately model the function. For example, the popular \COBYLA~\citep{Powell1994} method approximates the objective function using a local linear model.
Intuitively, while linear and quadratic surrogates are likely to be inadequate models globally, they can be accurate in a sufficiently small TR.
However, there are two challenges with traditional TR methods. First, deterministic examples such as \COBYLA are notorious for handling noisy observations poorly. Second, simple surrogate models might require overly small trust regions to provide accurate modeling behavior.
Therefore, we will use GP surrogate models within a TR. This allows us to inherit the robustness to noise and rigorous reasoning about uncertainty that global BO enjoys.

\paragraph{Trust regions.}
We choose our TR to be a hyperrectangle centered at the best solution found so far, denoted by $\xbest$.
In the noise-free case, we set $\xbest$ to the location of the best observation so far. In the presence of noise,
we use the observation with the smallest posterior mean under the surrogate model.
At the beginning of a given local optimization run, we initialize the \emph{base side length} of the TR to $\len \gets \len_{\textrm{init}}$.
The actual side length for each dimension is obtained from this base side length by rescaling according to its lengthscale $\lambda_i$ in the GP model while maintaining a total
volume of $\len^d$.
That is, $\len_i = \lambda_i L / (\prod_{j=1}^d \lambda_j)^{1/d}$.
To perform a single local optimization run, we utilize an acquisition function at each iteration $t$
to select a batch of~$\nbatch$ candidates~\smash{$\{\vec x_1^{(t)},\ldots,\vec x_{\nbatch}^{(t)}\}$}, restricted to be within the TR.
If $\len$ was large enough for the TR to contain the whole space, this would be equivalent to running standard global BO.
Therefore, the evolution of $\len$ is critical. On the one hand, a TR should be sufficiently
large to contain good solutions. On the other hand, it should be small enough to ensure that the local model is
accurate within the TR.
The typical behavior is to expand a TR when the optimizer ``makes progress'', i.e., it finds better solutions
in that region, and shrink it when the optimizer appears stuck. Therefore, following, e.g., \citet{nelder1965simplex}, we
will shrink a TR after too many consecutive ``failures'', and expand it after many consecutive ``successes''.
We define a ``success'' as a candidate that improves upon $\xbest$, and a ``failure'' as a candidate that does not.
After $\tau_{\text{succ}}$ consecutive successes, we double the size of the TR, i.e., $\len \gets \min\{\len_{\textrm{max}}, 2\len\}$.
After $\tau_{\text{fail}}$ consecutive failures, we halve the size of the TR: $\len \gets \len/2$. We reset the success and failure
counters to zero after we change the size of the TR.
Whenever~$\len$ falls below a given minimum threshold $\len_{\textrm{min}}$, we discard the respective TR and
initialize a new one with side length $\len_{\textrm{init}}$. Additionally, we do not let the side length expand to be larger than
a maximum threshold $\len_{\textrm{max}}$. Note that $\tau_{\text{succ}}$, $\tau_{\text{fail}}$, $\len_{\textrm{min}}$, $\len_{\textrm{max}}$,
and~$\len_{\textrm{init}}$ are hyperparameters of~\ALG; see the supplementary material for the values used in the experimental evaluation.

\paragraph{Trust region Bayesian optimization.}
So far, we have detailed a single \emph{local} BO strategy using a TR method. Intuitively, we could make this algorithm (more) global by random restarts.
However, from a probabilistic perspective, this is likely to utilize our evaluation budget inefficiently. Just as we reason about which candidates are
most promising within a local optimization run, we can reason about which local optimization run is ``most promising.''

Therefore, \ALG maintains~\narm trust regions \emph{simultaneously}.
Each trust region~$\text{TR}_\ell$ with~$\ell \in \{1, \ldots, \narm\}$ is a hyperrectangle of base side length~$\len_{\ell} \leq \len_{\textrm{max}}$,
and utilizes an independent local GP model.
This gives rise to a classical exploitation-exploration trade-off that we model by a multi-armed bandit that treats each TR as a lever.
Note that this provides an advantage over traditional TR algorithms in that \ALG puts a stronger emphasis on promising regions.

In each iteration, we need to select a batch of $\nbatch$ candidates drawn from the union of all trust regions, and update all
local optimization problems for which candidates were drawn. To solve this problem, we find that \TS
provides a principled solution to both the problem of selecting candidates within a single TR, and selecting
candidates across the set of trust regions simultaneously.
To select the $i$-th candidate from across the trust regions, we draw a realization of the posterior function from
the local GP within each TR: $f^{(i)}_{\ell} \sim \mathcal{GP}^{(t)}_{\ell}(\mu_{\ell}(\vec x), k_{\ell}(\vec
x, \vec x'))$, where $\mathcal{GP}^{(t)}_{\ell}$ is the GP posterior for $\text{TR}_{\ell}$ at iteration~$t$.
We then select the $i$-th candidate such that it minimizes the function value across all $\narm$ samples \emph{and} all trust regions:
\begin{equation*}
    \vec x_i^{(t)} \in \argmin_{\ell}\argmin_{\vec x \in \text{TR}_{\ell}} f^{(i)}_{\ell}\;\text{where}\;f^{(i)}_{\ell} \sim
    \mathcal{GP}^{(t)}_{\ell}(\mu_{\ell}(\vec x), k_{\ell}(\vec x, \vec x')).
\end{equation*}
That is, we select as point with the smallest function value after concatenating a Thompson sample from each TR for $i=1,\ldots,q$.
We refer to the supplementary material for additional details.
\begin{figure}[tb]
    \centering
    \label{fig:motivation}
    \includegraphics[width=0.97\textwidth]{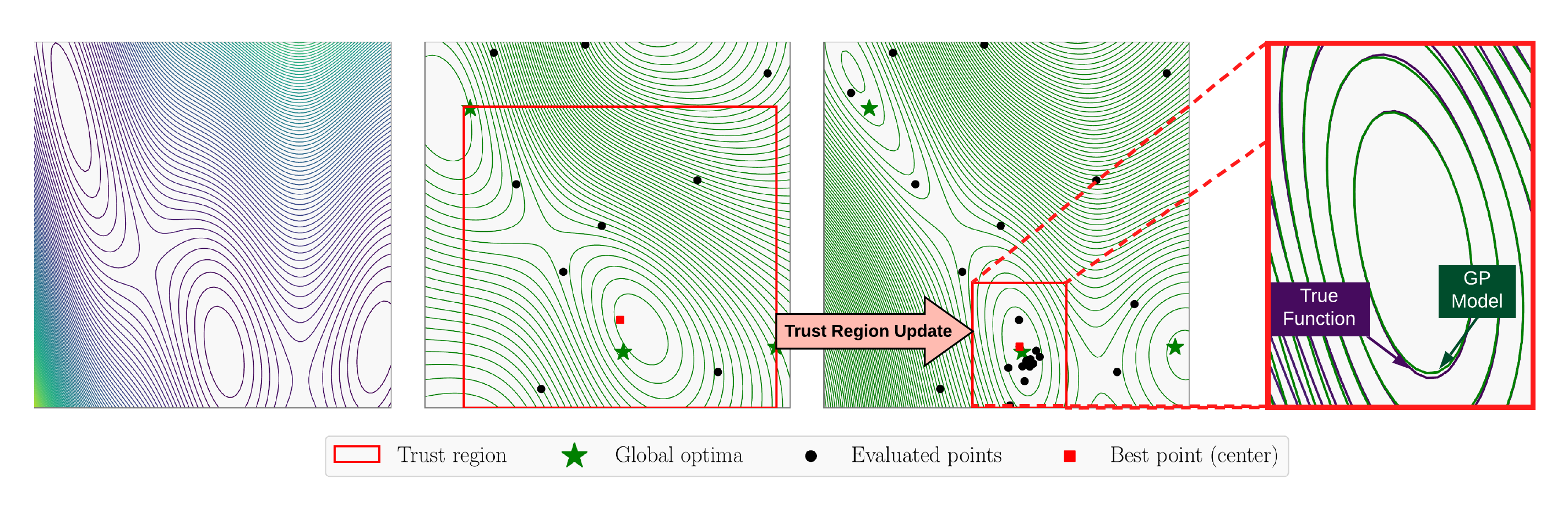}
    \caption{Illustration of the \ALG algorithm.
    \textbf{(Left)}~The true contours of the Branin function.
    \textbf{(Middle left)} The contours of the GP model fitted to the observations depicted by black dots. The current TR is shown as a red square. The global optima are indicated by the green stars.
    \textbf{(Middle right)} During the execution of the algorithm, the TR has moved towards the global optimum and has reduced in size. The area around the optimum has been sampled more densely in effect.
    \textbf{(Right)} The local GP model almost exactly fits the underlying function in the TR, despite having a poor global fit.
    }
\end{figure}

\section{Numerical experiments}
\label{sec:experiments}
In this section, we evaluate \ALG on a wide range of problems: a $14$D robot pushing problem, a $60$D rover trajectory planning problem, a $12$D cosmological
constant estimation problem, a $12$D lunar landing reinforcement learning problem, and a $200$D synthetic problem. All problems are multimodal and
challenging for many global optimization algorithms. We consider a variety of batch sizes and evaluation budgets to fully examine the performance
and robustness of \ALG. The values of $\tau_{\text{succ}}$, $\tau_{\text{fail}}$, $\len_{\textrm{min}}$, $\len_{\textrm{max}}$,
and~$\len_{\textrm{init}}$ are given in the supplementary material.

We compare \ALG to a comprehensive selection of state-of-the-art baselines:
\BFGS, \BOCK, \BOHAMIANN, \CMA, \BOBYQA, \EBO, \GPTS, \HeSBO, Nelder-Mead (\NM), and random search (\RS)\@.
Here, \GPTS refers to \TS with a global GP model using the Mat\'ern-$5/2$ kernel.
\HeSBO combines \GPTS with a subspace embedding and thus effectively optimizes in a low-dimensional space; this target dimension is set by the user.
Therefore, a small sample budget may suffice, which allows to run~$p$ invocations in parallel, following \citep{wang2016bayesian}.
This may improve the performance, since each embedding may "fail" with some probability \citep{HeSBO19}, i.e., it does not contain the active subspace even if it exists.
Note that \HeSBO-$p$ recommends a point of optimal posterior mean among the~$p$ GP-models; we use that point for the evaluation.
The standard acquisition criterion \texttt{EI} used in \BOCK and \BOHAMIANN is replaced by (batch) \TS, i.e., all methods use the same criterion which allows for a direct comparison.
Methods that attempt to learn an additive decomposition lack scalability and are thus omitted.
\BFGS approximates the gradient via finite differences and thus requires~$d{+}1$ evaluations for each step.
Furthermore, \NM, \BFGS, and \BOBYQA are inherently sequential and therefore have an edge by leveraging all gathered observations.
However, they are considerably more time consuming on a per-wall-time evaluation basis since we are working with large batches.

We supplement the optimization test problems with three additional experiments: i) one that shows that \ALG achieves a linear speed-up from
large batch sizes, ii) a comparison of local GPs and  global GPs on a control problem, and iii) an analytical experiment demonstrating the locality of \ALG.
Performance plots show the mean performances with one standard error.
Overall, we observe that \ALG consistently finds excellent solutions, outperforming the other methods on most problems.
Experimental results for a small budget experiment on four synthetic functions are shown in the supplement, where we also provide details on the experimental setup and runtimes for all algorithms.

\subsection{Robot pushing}
\label{sec:robo}
The robot pushing problem is a noisy $14$D control problem considered in \citet{wang2017batched}.
We run each method for a total of $10$K evaluations and batch size of $\nbatch = 50$. \ALG-$1$ and all other methods are initialized with $100$ points except for \ALG-$20$ where we use $50$ initial points for each trust region. This is to avoid having \ALG-$20$ consume its full evaluation budget on the initial points. We use \HeSBO-$5$ with target dimension~$8$.
\ALG-$\narm$ denotes the variant of~\ALG that maintains~$\narm$ local models in parallel.
Fig.~\ref{fig:robo_rover} shows the results:
\ALG-$1$ and \ALG-$20$ outperform the alternatives. \ALG-$20$ starts slower since it is initialized with $1$K points, but eventually outperforms \ALG-$1$.
\CMA and \BOBYQA outperform the other BO methods.
Note that \citet{wang2017batched} reported a median value of~$8.3$ for \EBO after $30$K evaluations, while \ALG-$1$ achieves a mean and median reward of around~$9.4$ after only $2$K samples.

\subsection{Rover trajectory planning}
\label{sec:rover}
\begin{figure}[htb]
    \centering
    \includegraphics[width=0.98\textwidth]{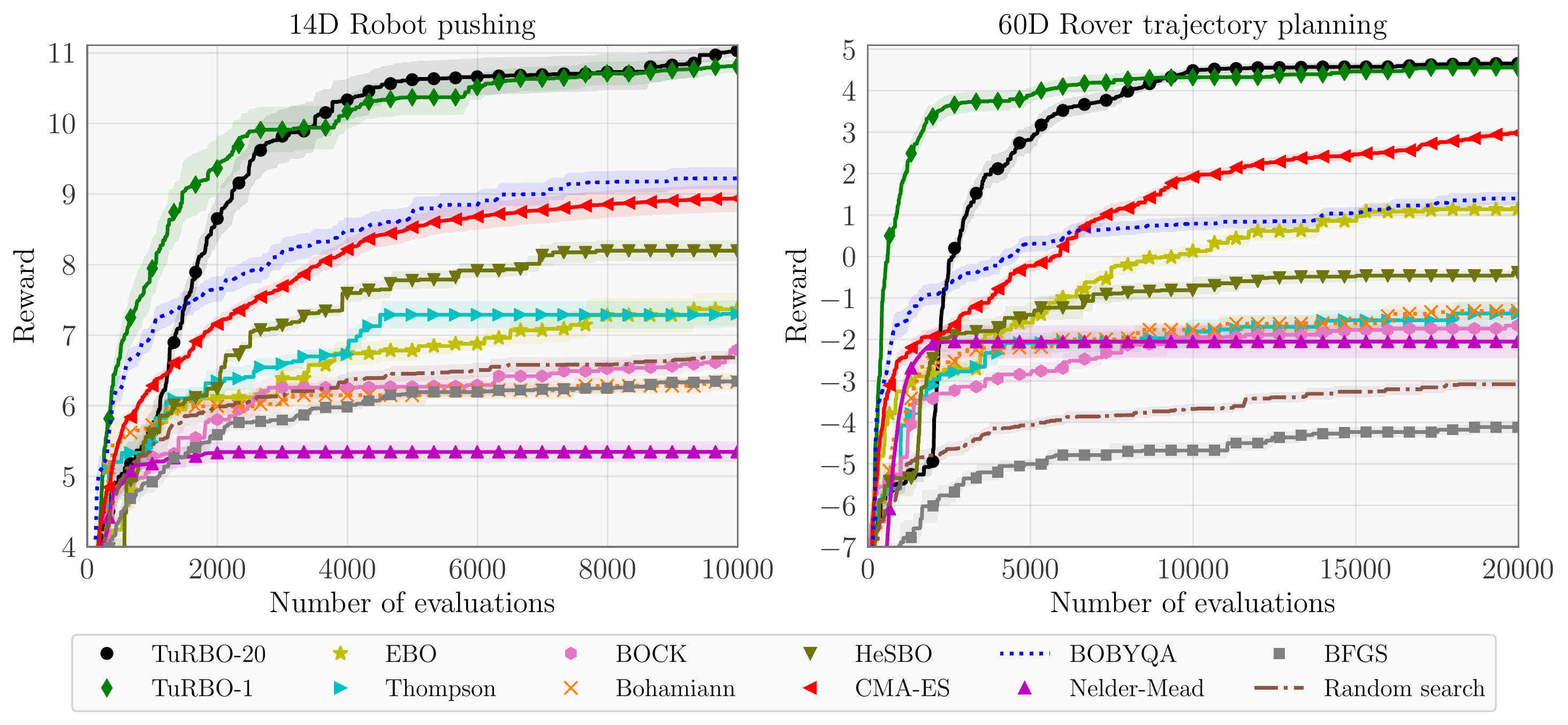}
    \caption{
        \textbf{14D Robot pushing (left):}~\ALG-$1$ and \ALG-$20$ perform well after a few thousand evaluations.
        \textbf{60D Rover trajectory planning (right):}~\ALG-$1$ and \ALG-$20$ achieve close to optimal objective values after $10$K evaluations. In both experiments \CMA and \BOBYQA are the runners up, and \HeSBO and \EBO perform best among the other BO methods.
    }
    \label{fig:robo_rover}
\end{figure}

Here the goal is to optimize the locations of $30$ points in the $2$D-plane that determine the trajectory of a rover~\citep{wang2017batched}.
Every algorithm is run for $200$ steps with a batch size of $\nbatch = 100$, thus collecting a total of $20$K evaluations.
We use $200$ initial points for all methods except for \ALG-$20$, where we use $100$ initial points for each region. Fig.~\ref{fig:robo_rover} summarizes the performance.
We observe that \ALG-$1$ and \ALG-$20$ outperform all other algorithms after a few thousand evaluations.
\ALG-$20$ once again starts slowly because of the initial $2$K random evaluations.
\citet{wang2017batched} reported a mean value of~$1.5$ for \EBO after $35$K evaluations, while \ALG-$1$ achieves a mean and median reward of about~$2$ after only $1$K evaluations. We use a target dimension of~$10$ for \HeSBO-$15$ in this experiment.

\subsection{Cosmological constant learning}
\label{sec:cosmo}
In the ``cosmological constants'' problem, the task is to calibrate a physics
simulator\footnote{\url{https://lambda.gsfc.nasa.gov/toolbox/lrgdr/}} to observed data.
The tunable parameters include various physical constants like the density of certain types of matter and Hubble's constant.
In this paper, we use a more challenging version of the problem in \citep{kandasamy2015high} by tuning $12$ parameters rather than $9$, and by using substantially larger parameter bounds. We used $2$K evaluations, a batch size of $\nbatch = 50$, and~$50$ initial points.
\ALG-$5$ uses $20$ initial points for each local model and \HeSBO-$4$ uses a target dimension of~$8$.
Fig.~\ref{fig:cosmo_lunar}~(left) shows the results, with \ALG-$5$ performing the best, followed by~\BOBYQA and \ALG-$1$.
\ALG-$1$ sometimes converges to a bad local optimum, which deteriorates the mean performance and demonstrates the importance of allocating samples across multiple trust regions.

\subsection{Lunar landing reinforcement learning}
\label{sec:lunar}
\begin{figure}[htb]
    \centering
    \includegraphics[width=0.99\textwidth]{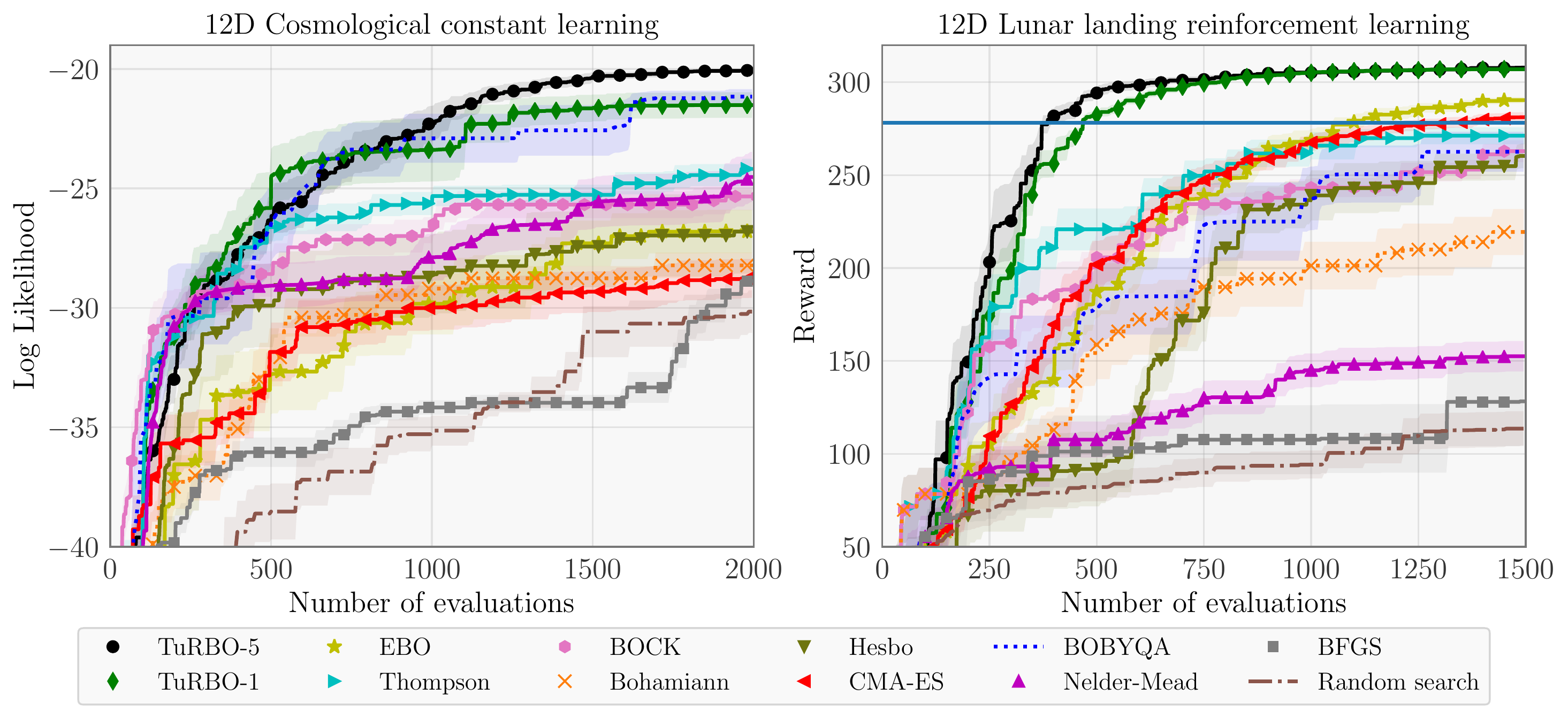}
    \caption{
        \textbf{12D Cosmological constant (left):}~\ALG-$5$ provides an improvement over \BOBYQA and \ALG-$1$. BO methods are distanced, with \TS performing best among them.
        \textbf{12D Lunar lander (right):}~\ALG-$5$, \ALG-$1$, \EBO, and \CMA learn better controllers than the original \OpenAI controller (solid blue horizontal line)\@.}
    \label{fig:cosmo_lunar}
    \raggedbottom
    \vspace{-2ex}
\end{figure}

Here the goal is to learn a controller for a lunar lander implemented in the OpenAI gym\footnote{\url{https://gym.openai.com/envs/LunarLander-v2}}.
The state space for the lunar lander is the position, angle, time derivatives, and whether or not either leg is in contact with the ground.
There are four possible action for each frame, each corresponding to firing a booster engine left, right, up, or doing nothing.
The objective is to maximize the average final reward over a fixed constant set of $50$ randomly generated terrains, initial positions, and velocities.
We observed that the simulation can be sensitive to even tiny perturbations.
Fig.~\ref{fig:cosmo_lunar} shows the results for a total of $1500$ function evaluations, batch size $\nbatch = 50$, and $50$ initial points for all algorithms except for \ALG-$5$ which uses 20 initial points for each local region. For this problem, we use \HeSBO-$3$ in an $8$-dimensional subspace.
\ALG-$5$ and \ALG-$1$ learn the best controllers; and in particular achieves better rewards than the handcrafted controller provided by OpenAI whose performance is depicted by the blue horizontal line.

\subsection{The 200-dimensional Ackley function}
\label{sec:200d}
\begin{figure}[H]
    \centering
    \includegraphics[width=0.6\textwidth]{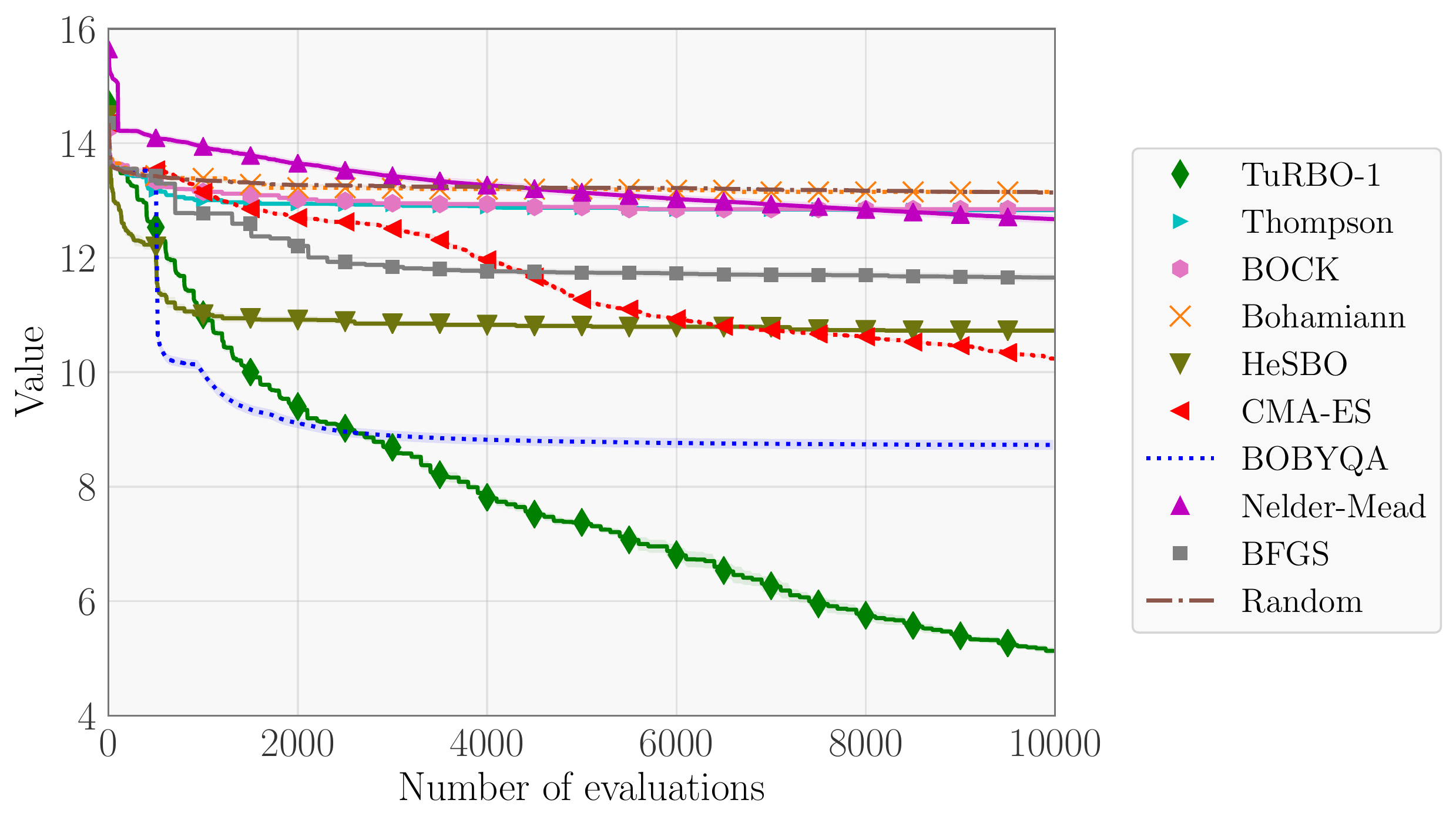}
    \caption{
        \textbf{200D Ackley function:}~\ALG-$1$ clearly outperforms the other baselines. \BOBYQA makes good initial progress but consistently converges to sub-optimal local minima.
    }
    \label{fig:ackley200}
\end{figure}

We examine performances on the $200$-dimensional Ackley function in the domain \smash{$[-5, 10]^{200}$}.
We only consider \ALG-$1$ because of the large number of dimensions where there may not be a benefit from using multiple TRs. EBO is excluded from the plot since its computation time exceeded $30$ days per replication. \HeSBO-$5$ uses a target dimension of~$20$.
Fig.~\ref{fig:ackley200} shows the results for a total of 10K function evaluations, batch size $\nbatch = 100$, and $200$ initial points for all algorithms.

\HeSBO-$5$, with a target dimension of $20$, and \BOBYQA perform well initially, but are eventually outperformed by \ALG-$1$ that achieves the best solutions.
The good performance of~\HeSBO is particularly interesting, since this benchmark has no redundant dimensions and thus should be challenging for that embedding-based approach.
This confirms similar findings in~\citep{HeSBO19}.
BO methods that use a global GP model over-emphasize exploration and make little progress.

\subsection{The advantage of local models over global models}
\label{sec:local_vs_global}
\begin{figure}[htbp]
    \centering
    \includegraphics[width=0.99\textwidth]{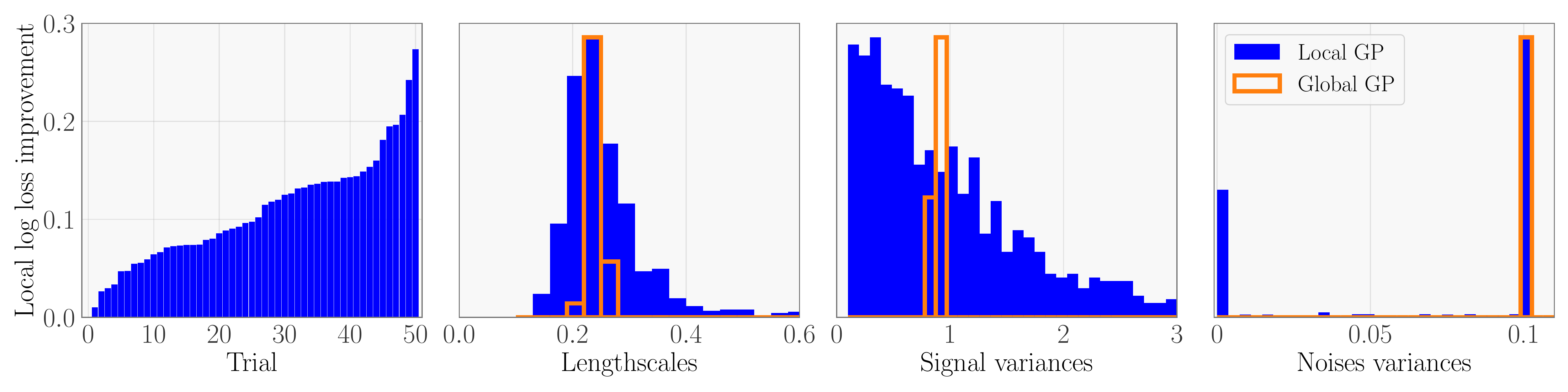}
    \caption{
        \textbf{Local and global GPs on log loss (left):}~We show the improvement in test set log loss (nats/test point) of the local model over the global model by repeated trial.
        The local GP increases in performance in every trial.
        Trials are sorted in order of performance gain.
        This shows a substantial mean improvement of $0.110$ nats.
        \textbf{Learned hypers (right three figures):}~A histogram plot of the hyperparameters learned by the local (blue) and global (orange) GPs pooled across all repeated trials.
        The local GPs show a much wider range of hyperparameters that can specialize per region.
    }
    \label{fig:local_vs_global}
\end{figure}

We investigate the performance of local and global GP models on the $14$D robot pushing problem from Sect.~\ref{sec:robo}.
We replicate the conditions from the optimization experiments as closely as possible for a regression experiment, including for example parameter bounds.
We choose~$20$ uniformly distributed hypercubes of (base) side length~$0.4$, each containing~$200$ uniformly distributed training points.
We train a global GP on all $4000$ samples, as well as a separate local GP for each hypercube.
For the sake of illustration, we used an isotropic kernel for these experiments.
The local GPs have the advantage of being able to learn different hyperparameters in each region while the global GP has the advantage of having access to all of the data.
Fig.~\ref{fig:local_vs_global} shows the predictive performance (in log loss) on held-out data.
We also show the distribution of fitted hyperparameters for both the local and global GPs.
We see that the hyperparameters (especially the signal variance) vary substantially across regions.
Furthermore, the local GPs perform better than the global GP in every repeated trial.
The global model has an average log loss of~$1.284$ while the local model has an average log loss of~$1.174$ across $50$ trials; the improvement is significant under a $t$-test at \smash{$p < 10^{-4}$}.
This experiment confirms that we improve the predictive power of the models and also reduce the computational overhead of the GP by using the local approach.
The learned local noise variance in Fig.~\ref{fig:local_vs_global} is bimodal, confirming the heteroscedasticity in the objective across regions.
The global GP is required to learn the high noise value to avoid a penalty for outliers.

\subsection{Why high-dimensional spaces are challenging}
In this section, we illustrate why the restarting and banditing strategy of \ALG is so effective.
Each TR restart finds distant solutions of varying quality, which highlights the multimodal nature of the problem.
This gives \ALG-$\narm$ a distinct advantage.

\begin{figure}[htb]
    \centering
    \includegraphics[width=0.99\textwidth]{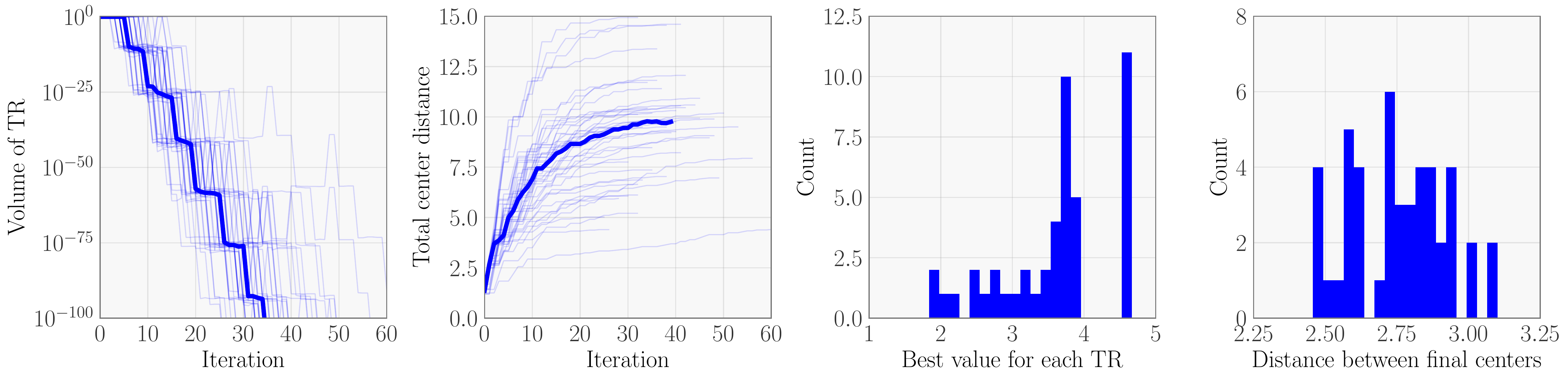}
    \caption{
        Performance statistics for $50$ restarts of \ALG-$1$ on the $60$D rover trajectory planning problem.
        The domain is scaled to \smash{$[0, 1]^{60}$}.
        \textbf{Trust region volume (left):}~We see that the volume of the TR decreases with the iterations. Each TR is shown by a light blue line, and their average in solid blue.
        \textbf{Total center distance (middle left):}~The cumulative Euclidean distance that each TR center has moved (trajectory arc length). This confirms the balance between initial exploration and final exploitation.
        \textbf{Best value found (middle right):}~The best function value found during each run of \ALG-$1$. The solutions vary in quality, which explains why our bandit approach works well.
        \textbf{Distance between final TR centers (right):}~Minimum distances between final TR centers, which shows that each restart leads to a different part of the space.
    }
    \label{fig:tr_details}
\end{figure}

We ran \ALG-$1$ (with a single trust region) for $50$ restarts on the $60$D rover trajectory planning problem from Sect.~\ref{sec:rover} and logged the volume of the TR and its center after each iteration.
Fig.~\ref{fig:tr_details} shows the volume of the TR, the arclength of the TR center's trajectory, the final objective value, and the distance each final solution has to its nearest neighbor.
The left two plots confirm that, within a trust region, the optimization is indeed highly local. The volume of any given trust region decreases rapidly and is only a small fraction of the total search space. From the two plots on the right, we see that the solutions found by \ALG are far apart with varying quality, demonstrating the value of performing multiple local search runs in parallel.

\subsection{The efficiency of large batches}
\label{sec:batch}
\begin{figure}[htb]
    \centering
    \includegraphics[width=0.99\textwidth]{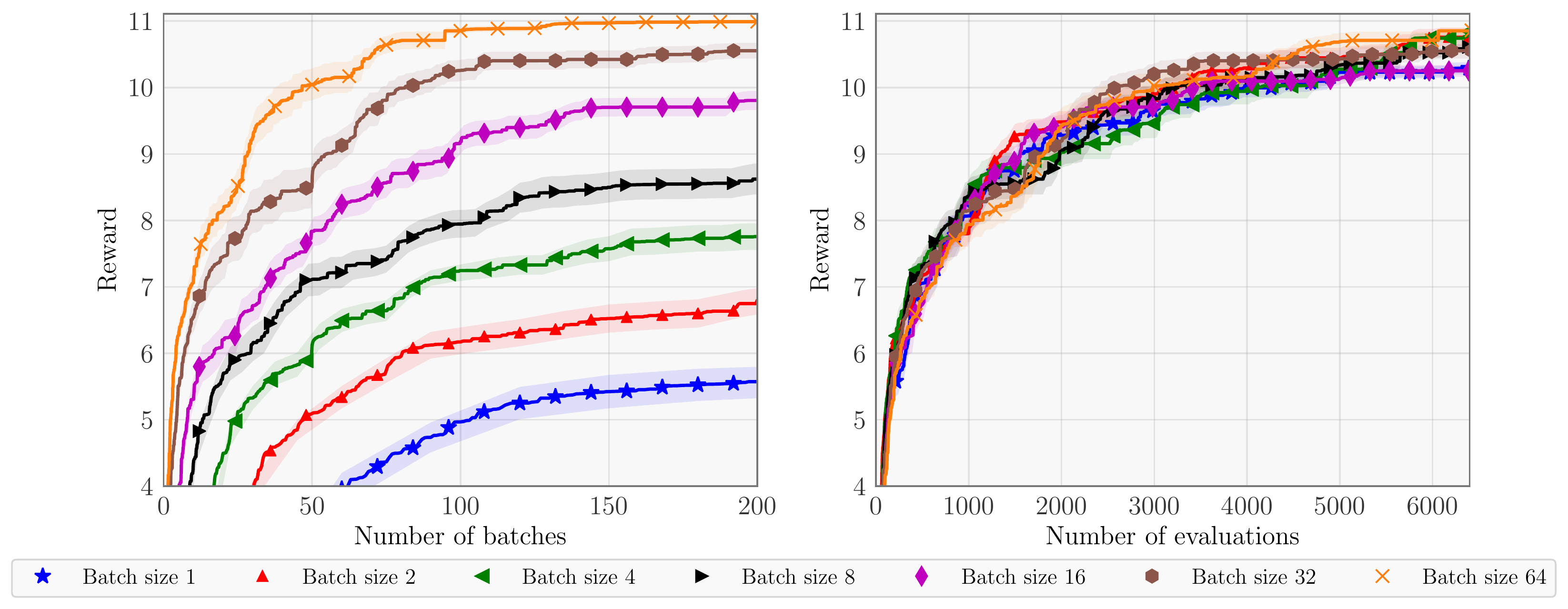}
    \caption{
        We evaluate \ALG for different batch sizes. On the left, we see that larger batches provide better solutions at the same number of steps.  On the right, we see that this reduction in wall-clock time does not come at the expense of efficacy, with large batches providing nearly linear speed up.
    }
    \label{fig:batch_robopush}
    \vspace{-2ex}
\end{figure}

Recall that combining multiple samples into single batches provides substantial speed-ups in terms of wall-clock time but poses the risk of inefficiencies since sequential sampling has the advantage of leveraging more information.
In this section, we investigate whether large batches are efficient for \ALG.
Note that \citet{hernandez2017parallel} and \citet{kandasamy2018parallelised} have shown that the \TS acquisition function is efficient for batch acquisition with a single global surrogate model.
We study \ALG-$1$ on the robot pushing problem from Sect.~\ref{sec:robo} with batch sizes $\nbatch \in \{1, 2, 4, \ldots, 64\}$.
The algorithm takes $\max\{200 \nbatch, 6400\}$ samples for each batch size and we average the results over $30$ replications.
Fig.~\ref{fig:batch_robopush}~(left) shows the reward for each batch size with respect to the number of batches: we see that larger batch sizes obtain better results for the same number of iterations.
Fig.~\ref{fig:batch_robopush}~(right) shows the performance as a function of evaluations.  We see that the speed-up is essentially linear.

\section{Conclusions}
\label{sec:conclusions}
The global optimization of computationally expensive black-box functions in high-dimensional spaces is an important and timely topic~\citep{frazier2018tutorial,HeSBO19}.
We proposed the \ALG algorithm which takes a novel local approach to global optimization.
Instead of fitting a global surrogate model and trading off exploration and exploitation on the whole search space, \ALG maintains a collection of local probabilistic models. These models provide local search trajectories that are able to quickly discover excellent objective values.
This local approach is complemented with a global bandit strategy that allocates samples across these trust regions, implicitly trading off exploration and exploitation.
A comprehensive experimental evaluation demonstrates that \ALG outperforms the state-of-the-art Bayesian optimization and operations research methods on a variety of real-world complex tasks.

In the future, we plan on extending \ALG to learn local low-dimensional structure to improve the accuracy of the local Gaussian process model.
This extension is particularly interesting in high-dimensional optimization when derivative information is available~\citep{constantine2015active,eriksson2018scaling,wu2017bayesian}.
This situation often arises in engineering, where objectives are often modeled by PDEs solved by adjoint methods, and in machine learning where gradients are available via automated differentiation. Ultimately, it is our hope that this work spurs interest in the merits of Bayesian \emph{local} optimization, particularly in the high-dimensional setting.


\bibliographystyle{abbrvnat}
\bibliography{bib_trts}

\begin{thebibliography}{58}
\providecommand{\natexlab}[1]{#1}
\providecommand{\url}[1]{\texttt{#1}}
\expandafter\ifx\csname urlstyle\endcsname\relax
  \providecommand{\doi}[1]{doi: #1}\else
  \providecommand{\doi}{doi: \begingroup \urlstyle{rm}\Url}\fi

\bibitem[Acerbi and Ji(2017)]{acerbi2017practical}
L.~Acerbi and W.~Ji.
\newblock Practical {Bayesian} optimization for model fitting with {Bayesian}
  adaptive direct search.
\newblock In \emph{Advances in neural information processing systems}, pages
  1836--1846, 2017.

\bibitem[Akrour et~al.(2017)Akrour, Sorokin, Peters, and
  Neumann]{akrour2017local}
R.~Akrour, D.~Sorokin, J.~Peters, and G.~Neumann.
\newblock Local {Bayesian} optimization of motor skills.
\newblock In \emph{Proceedings of the 34th International Conference on Machine
  Learning-Volume 70}, pages 41--50. JMLR. org, 2017.

\bibitem[Assael et~al.(2014)Assael, Wang, Shahriari, and
  de~Freitas]{assael2014heteroscedastic}
J.-A.~M. Assael, Z.~Wang, B.~Shahriari, and N.~de~Freitas.
\newblock Heteroscedastic treed {B}ayesian optimisation.
\newblock \emph{arXiv preprint arXiv:1410.7172}, 2014.

\bibitem[Baptista and Poloczek(2018)]{baptista2018bayesian}
R.~Baptista and M.~Poloczek.
\newblock Bayesian optimization of combinatorial structures.
\newblock In \emph{International Conference on Machine Learning}, pages
  462--471, 2018.

\bibitem[Binois et~al.(2015)Binois, Ginsbourger, and
  Roustant]{binois2015warped}
M.~Binois, D.~Ginsbourger, and O.~Roustant.
\newblock A warped kernel improving robustness in {Bayesian} optimization via
  random embeddings.
\newblock In \emph{International Conference on Learning and Intelligent
  Optimization}, pages 281--286. Springer, 2015.

\bibitem[Binois et~al.(2019 (to appear))Binois, Ginsbourger, and
  Roustant]{binois2017choice}
M.~Binois, D.~Ginsbourger, and O.~Roustant.
\newblock On the choice of the low-dimensional domain for global optimization
  via random embeddings.
\newblock \emph{Journal of Global Optimization}, 2019 (to appear).

\bibitem[Calandra et~al.(2016)Calandra, Seyfarth, Peters, and
  Deisenroth]{calandra2016bayesian}
R.~Calandra, A.~Seyfarth, J.~Peters, and M.~P. Deisenroth.
\newblock Bayesian optimization for learning gaits under uncertainty.
\newblock \emph{Annals of Mathematics and Artificial Intelligence}, 76\penalty0
  (1-2):\penalty0 5--23, 2016.

\bibitem[Chen et~al.(2012)Chen, Castro, and Krause]{chen2012joint}
B.~Chen, R.~M. Castro, and A.~Krause.
\newblock Joint optimization and variable selection of high-dimensional
  {G}aussian processes.
\newblock In \emph{Proceedings of the International Conference on Machine
  Learning}, pages 1379--1386. Omnipress, 2012.

\bibitem[Chevalier and Ginsbourger(2013)]{chevalier2013fast}
C.~Chevalier and D.~Ginsbourger.
\newblock Fast computation of the multi-points expected improvement with
  applications in batch selection.
\newblock In \emph{International Conference on Learning and Intelligent
  Optimization}, pages 59--69. Springer, 2013.

\bibitem[Constantine(2015)]{constantine2015active}
P.~G. Constantine.
\newblock \emph{Active subspaces: Emerging ideas for dimension reduction in
  parameter studies}, volume~2.
\newblock SIAM, 2015.

\bibitem[Dong et~al.(2017{\natexlab{a}})Dong, Eriksson, Nickisch, Bindel, and
  Wilson]{dong2017scalable}
K.~Dong, D.~Eriksson, H.~Nickisch, D.~Bindel, and A.~G. Wilson.
\newblock Scalable log determinants for {Gaussian} process kernel learning.
\newblock In \emph{Advances in Neural Information Processing Systems}, pages
  6327--6337, 2017{\natexlab{a}}.

\bibitem[Dong et~al.(2017{\natexlab{b}})Dong, Eckman, Zhao, Henderson, and
  Poloczek]{dong2017empirically}
N.~A. Dong, D.~J. Eckman, X.~Zhao, S.~G. Henderson, and M.~Poloczek.
\newblock Empirically comparing the finite-time performance of
  simulation-optimization algorithms.
\newblock In \emph{Winter Simulation Conference}, pages 2206--2217. IEEE,
  2017{\natexlab{b}}.

\bibitem[Eriksson et~al.(2018)Eriksson, Dong, Lee, Bindel, and
  Wilson]{eriksson2018scaling}
D.~Eriksson, K.~Dong, E.~Lee, D.~Bindel, and A.~G. Wilson.
\newblock Scaling {Gaussian} process regression with derivatives.
\newblock In \emph{Advances in Neural Information Processing Systems}, pages
  6867--6877, 2018.

\bibitem[Frazier(2018)]{frazier2018tutorial}
P.~I. Frazier.
\newblock A tutorial on {Bayesian} optimization.
\newblock \emph{arXiv preprint arXiv:1807.02811}, 2018.

\bibitem[Gardner et~al.(2017)Gardner, Guo, Weinberger, Garnett, and
  Grosse]{gardner2017discovering}
J.~Gardner, C.~Guo, K.~Weinberger, R.~Garnett, and R.~Grosse.
\newblock Discovering and exploiting additive structure for {Bayesian}
  optimization.
\newblock In \emph{International Conference on Artificial Intelligence and
  Statistics}, pages 1311--1319, 2017.

\bibitem[Gardner et~al.(2018)Gardner, Pleiss, Weinberger, Bindel, and
  Wilson]{gardner2018gpytorch}
J.~Gardner, G.~Pleiss, K.~Q. Weinberger, D.~Bindel, and A.~G. Wilson.
\newblock {GPyTorch}: {Blackbox} matrix-matrix {Gaussian} process inference
  with {GPU} acceleration.
\newblock In \emph{Advances in Neural Information Processing Systems}, pages
  7576--7586, 2018.

\bibitem[Garnett et~al.(2014)Garnett, Osborne, and Hennig]{garnett2013active}
R.~Garnett, M.~Osborne, and P.~Hennig.
\newblock Active learning of linear embeddings for {Gaussian} processes.
\newblock In \emph{30th Conference on Uncertainty in Artificial Intelligence
  (UAI 2014)}, pages 230--239, 2014.

\bibitem[Gonz{\'a}lez et~al.(2016)Gonz{\'a}lez, Dai, Hennig, and
  Lawrence]{gonzalez2016batch}
J.~Gonz{\'a}lez, Z.~Dai, P.~Hennig, and N.~Lawrence.
\newblock Batch {B}ayesian optimization via local penalization.
\newblock In \emph{Artificial intelligence and statistics}, pages 648--657,
  2016.

\bibitem[Hansen(2006)]{hansen2006cma}
N.~Hansen.
\newblock The {CMA} evolution strategy: {A} comparing review.
\newblock In \emph{Towards a New Evolutionary Computation}, pages 75--102.
  Springer, 2006.

\bibitem[Hern{\'a}ndez-Lobato et~al.(2017)Hern{\'a}ndez-Lobato, Requeima,
  Pyzer-Knapp, and Aspuru-Guzik]{hernandez2017parallel}
J.~M. Hern{\'a}ndez-Lobato, J.~Requeima, E.~O. Pyzer-Knapp, and
  A.~Aspuru-Guzik.
\newblock Parallel and distributed {Thompson} sampling for large-scale
  accelerated exploration of chemical space.
\newblock In \emph{Proceedings of the International Conference on Machine
  Learning}, pages 1470--1479, 2017.

\bibitem[Hutter et~al.(2011)Hutter, Hoos, and
  Leyton-Brown]{hutter2011sequential}
F.~Hutter, H.~H. Hoos, and K.~Leyton-Brown.
\newblock Sequential model-based optimization for general algorithm
  configuration.
\newblock In \emph{International Conference on Learning and Intelligent
  Optimization}, pages 507--523. Springer, 2011.

\bibitem[Jin et~al.(2005)Jin, Branke, et~al.]{jin2005evolutionary}
Y.~Jin, J.~Branke, et~al.
\newblock Evolutionary optimization in uncertain environments-a survey.
\newblock \emph{IEEE Transactions on Evolutionary Computation}, 9\penalty0
  (3):\penalty0 303--317, 2005.

\bibitem[Johnson(2014)]{johnson2014nlopt}
S.~G. Johnson.
\newblock The nlopt nonlinear-optimization package, 2014.
\newblock \emph{URL: http://ab-initio.mit.edu/nlopt}, 2014.

\bibitem[Jones et~al.(2014)Jones, Oliphant, and Peterson]{jones2014scipy}
E.~Jones, T.~Oliphant, and P.~Peterson.
\newblock {SciPy}: {Open} source scientific tools for {Python}.
\newblock 2014.

\bibitem[Kandasamy et~al.(2015)Kandasamy, Schneider, and
  P{\'o}czos]{kandasamy2015high}
K.~Kandasamy, J.~Schneider, and B.~P{\'o}czos.
\newblock High dimensional {Bayesian} optimisation and bandits via additive
  models.
\newblock In \emph{International Conference on Machine Learning}, pages
  295--304, 2015.

\bibitem[Kandasamy et~al.(2018)Kandasamy, Krishnamurthy, Schneider, and
  P{\'o}czos]{kandasamy2018parallelised}
K.~Kandasamy, A.~Krishnamurthy, J.~Schneider, and B.~P{\'o}czos.
\newblock Parallelised {Bayesian} optimisation via {Thompson} sampling.
\newblock In \emph{International Conference on Artificial Intelligence and
  Statistics}, pages 133--142, 2018.

\bibitem[Krityakierne and Ginsbourger(2015)]{krityakierne2015global}
T.~Krityakierne and D.~Ginsbourger.
\newblock Global optimization with sparse and local {Gaussian} process models.
\newblock In \emph{International Workshop on Machine Learning, Optimization and
  Big Data}, pages 185--196. Springer, 2015.

\bibitem[Marmin et~al.(2015)Marmin, Chevalier, and
  Ginsbourger]{marmin2015differentiating}
S.~Marmin, C.~Chevalier, and D.~Ginsbourger.
\newblock Differentiating the multipoint expected improvement for optimal batch
  design.
\newblock In \emph{International Workshop on Machine Learning, Optimization and
  Big Data}, pages 37--48. Springer, 2015.

\bibitem[McKay et~al.(1979)McKay, Beckman, and Conover]{McKay1979}
M.~D. McKay, R.~J. Beckman, and W.~J. Conover.
\newblock Comparison of three methods for selecting values of input variables
  in the analysis of output from a computer code.
\newblock \emph{Technometrics}, 21\penalty0 (2):\penalty0 239--245, 1979.

\bibitem[McLeod et~al.(2018)McLeod, Roberts, and
  Osborne]{mcleod2018optimization}
M.~McLeod, S.~Roberts, and M.~A. Osborne.
\newblock Optimization, fast and slow: Optimally switching between local and
  {Bayesian} optimization.
\newblock In \emph{International Conference on Machine Learning}, pages
  3440--3449, 2018.

\bibitem[Mutny and Krause(2018)]{mutny2018efficient}
M.~Mutny and A.~Krause.
\newblock Efficient high dimensional {Bayesian} optimization with additivity
  and quadrature {Fourier} features.
\newblock In \emph{Advances in Neural Information Processing Systems}, pages
  9005--9016, 2018.

\bibitem[Nayebi et~al.(2019)Nayebi, Munteanu, and Poloczek]{HeSBO19}
A.~Nayebi, A.~Munteanu, and M.~Poloczek.
\newblock A framework for bayesian optimization in embedded subspaces.
\newblock In \emph{International Conference on Machine Learning}, pages
  4752--4761, 2019.
\newblock The code is available at https://github.com/aminnayebi/HesBO.

\bibitem[Nelder and Mead(1965)]{nelder1965simplex}
J.~A. Nelder and R.~Mead.
\newblock A simplex method for function minimization.
\newblock \emph{The Computer Journal}, 7\penalty0 (4):\penalty0 308--313, 1965.

\bibitem[Oh et~al.(2018)Oh, Gavves, and Welling]{oh2018bock}
C.~Oh, E.~Gavves, and M.~Welling.
\newblock {BOCK} : {B}ayesian optimization with cylindrical kernels.
\newblock In \emph{Proceedings of the 35th International Conference on Machine
  Learning}, volume~80, pages 3868--3877, 2018.

\bibitem[Powell(1994)]{Powell1994}
M.~J. Powell.
\newblock A direct search optimization method that models the objective and
  constraint functions by linear interpolation.
\newblock In \emph{Advances in Optimization and Numerical Analysis}, pages
  51--67. Springer, 1994.

\bibitem[Powell(2007)]{powell2007view}
M.~J. Powell.
\newblock A view of algorithms for optimization without derivatives.
\newblock \emph{Mathematics Today-Bulletin of the Institute of Mathematics and
  its Applications}, 43\penalty0 (5):\penalty0 170--174, 2007.

\bibitem[Regis and Shoemaker(2007)]{regis2007stochastic}
R.~G. Regis and C.~A. Shoemaker.
\newblock A stochastic radial basis function method for the global optimization
  of expensive functions.
\newblock \emph{INFORMS Journal on Computing}, 19\penalty0 (4):\penalty0
  497--509, 2007.

\bibitem[Regis and Shoemaker(2013)]{regis2013combining}
R.~G. Regis and C.~A. Shoemaker.
\newblock Combining radial basis function surrogates and dynamic coordinate
  search in high-dimensional expensive black-box optimization.
\newblock \emph{Engineering Optimization}, 45\penalty0 (5):\penalty0 529--555,
  2013.

\bibitem[Rolland et~al.(2018)Rolland, Scarlett, Bogunovic, and
  Cevher]{rolland2018high}
P.~Rolland, J.~Scarlett, I.~Bogunovic, and V.~Cevher.
\newblock High-dimensional {B}ayesian optimization via additive models with
  overlapping groups.
\newblock In \emph{International Conference on Artificial Intelligence and
  Statistics}, pages 298--307, 2018.

\bibitem[Russo et~al.(2018)Russo, Van~Roy, Kazerouni, Osband, Wen,
  et~al.]{russo2018tutorial}
D.~J. Russo, B.~Van~Roy, A.~Kazerouni, I.~Osband, Z.~Wen, et~al.
\newblock A tutorial on {Thompson} sampling.
\newblock \emph{Foundations and Trends in Machine Learning}, 11\penalty0
  (1):\penalty0 1--96, 2018.

\bibitem[Shah and Ghahramani(2015)]{shah2015parallel}
A.~Shah and Z.~Ghahramani.
\newblock Parallel predictive entropy search for batch global optimization of
  expensive objective functions.
\newblock In \emph{Advances in Neural Information Processing Systems}, pages
  3330--3338, 2015.

\bibitem[Shahriari et~al.(2016)Shahriari, Swersky, Wang, Adams, and
  De~Freitas]{shahriari2016taking}
B.~Shahriari, K.~Swersky, Z.~Wang, R.~P. Adams, and N.~De~Freitas.
\newblock Taking the human out of the loop: {A} review of {Bayesian}
  optimization.
\newblock \emph{Proceedings of the IEEE}, 104\penalty0 (1):\penalty0 148--175,
  2016.

\bibitem[Snoek et~al.(2012)Snoek, Larochelle, and Adams]{snoek2012practical}
J.~Snoek, H.~Larochelle, and R.~P. Adams.
\newblock Practical {Bayesian} optimization of machine learning algorithms.
\newblock In \emph{Advances in Neural Information Processing Systems}, pages
  2951--2959, 2012.

\bibitem[Snoek et~al.(2014)Snoek, Swersky, Zemel, and Adams]{snoek2014input}
J.~Snoek, K.~Swersky, R.~Zemel, and R.~Adams.
\newblock Input warping for {Bayesian} optimization of non-stationary
  functions.
\newblock In \emph{International Conference on Machine Learning}, pages
  1674--1682, 2014.

\bibitem[Snoek et~al.(2015)Snoek, Rippel, Swersky, Kiros, Satish, Sundaram,
  Patwary, Prabhat, and Adams]{snoek2015scalable}
J.~Snoek, O.~Rippel, K.~Swersky, R.~Kiros, N.~Satish, N.~Sundaram, M.~Patwary,
  M.~Prabhat, and R.~Adams.
\newblock Scalable {Bayesian} optimization using deep neural networks.
\newblock In \emph{International Conference on Machine Learning}, pages
  2171--2180, 2015.

\bibitem[Springenberg et~al.(2016)Springenberg, Klein, Falkner, and
  Hutter]{springenberg2016bayesian}
J.~T. Springenberg, A.~Klein, S.~Falkner, and F.~Hutter.
\newblock Bayesian optimization with robust {Bayesian} neural networks.
\newblock In \emph{Advances in Neural Information Processing Systems}, pages
  4134--4142, 2016.

\bibitem[Taddy et~al.(2009)Taddy, Lee, Gray, and Griffin]{taddy2009bayesian}
M.~A. Taddy, H.~K. Lee, G.~A. Gray, and J.~D. Griffin.
\newblock Bayesian guided pattern search for robust local optimization.
\newblock \emph{Technometrics}, 51\penalty0 (4):\penalty0 389--401, 2009.

\bibitem[Tegmark et~al.(2006)Tegmark, Eisenstein, Strauss, Weinberg, Blanton,
  Frieman, Fukugita, Gunn, Hamilton, Knapp, et~al.]{tegmark2006cosmological}
M.~Tegmark, D.~J. Eisenstein, M.~A. Strauss, D.~H. Weinberg, M.~R. Blanton,
  J.~A. Frieman, M.~Fukugita, J.~E. Gunn, A.~J. Hamilton, G.~R. Knapp, et~al.
\newblock Cosmological constraints from the {SDSS} luminous red galaxies.
\newblock \emph{Physical Review D}, 74\penalty0 (12):\penalty0 123507, 2006.

\bibitem[Thompson(1933)]{thompson1933likelihood}
W.~R. Thompson.
\newblock On the likelihood that one unknown probability exceeds another in
  view of the evidence of two samples.
\newblock \emph{Biometrika}, 25\penalty0 (3/4):\penalty0 285--294, 1933.

\bibitem[Wabersich and Toussaint(2016)]{wabersich2016advancing}
K.~P. Wabersich and M.~Toussaint.
\newblock Advancing {Bayesian} optimization: The mixed-global-local {(MGL)}
  kernel and length-scale cool down.
\newblock \emph{arXiv preprint arXiv:1612.03117}, 2016.

\bibitem[Wang et~al.(2016{\natexlab{a}})Wang, Clark, Liu, and
  Frazier]{wang2016parallel}
J.~Wang, S.~C. Clark, E.~Liu, and P.~I. Frazier.
\newblock Parallel {B}ayesian global optimization of expensive functions.
\newblock \emph{arXiv preprint arXiv:1602.05149}, 2016{\natexlab{a}}.

\bibitem[Wang et~al.(2016{\natexlab{b}})Wang, Hutter, Zoghi, Matheson, and
  de~Freitas]{wang2016bayesian}
Z.~Wang, F.~Hutter, M.~Zoghi, D.~Matheson, and N.~de~Freitas.
\newblock Bayesian optimization in a billion dimensions via random embeddings.
\newblock \emph{Journal of Artificial Intelligence Research}, 55:\penalty0
  361--387, 2016{\natexlab{b}}.

\bibitem[Wang et~al.(2018)Wang, Gehring, Kohli, and Jegelka]{wang2017batched}
Z.~Wang, C.~Gehring, P.~Kohli, and S.~Jegelka.
\newblock Batched large-scale {Bayesian} optimization in high-dimensional
  spaces.
\newblock In \emph{International Conference on Artificial Intelligence and
  Statistics}, pages 745--754, 2018.

\bibitem[Williams and Rasmussen(2006)]{williams2006gaussian}
C.~K. Williams and C.~E. Rasmussen.
\newblock \emph{Gaussian processes for machine learning}, volume~2.
\newblock MIT Press, 2006.

\bibitem[Wu and Frazier(2016)]{wu2016parallel}
J.~Wu and P.~Frazier.
\newblock The parallel knowledge gradient method for batch {Bayesian}
  optimization.
\newblock In \emph{Advances in Neural Information Processing Systems}, pages
  3126--3134, 2016.

\bibitem[Wu et~al.(2017)Wu, Poloczek, Wilson, and Frazier]{wu2017bayesian}
J.~Wu, M.~Poloczek, A.~G. Wilson, and P.~Frazier.
\newblock Bayesian optimization with gradients.
\newblock In \emph{Advances in Neural Information Processing Systems}, pages
  5267--5278, 2017.

\bibitem[Yuan(2000)]{Yuan2000}
Y.~Yuan.
\newblock A review of trust region algorithms for optimization.
\newblock In \emph{International Council for Industrial and Applied
  Mathematics}, volume~99, pages 271--282, 2000.

\bibitem[Zhu et~al.(1997)Zhu, Byrd, Lu, and Nocedal]{Zhu1997}
C.~Zhu, R.~H. Byrd, P.~Lu, and J.~Nocedal.
\newblock Algorithm 778: {L-BFGS-B}: {Fortran} subroutines for large-scale
  bound-constrained optimization.
\newblock \emph{ACM Transactions on Mathematical Software (TOMS)}, 23\penalty0
  (4):\penalty0 550--560, 1997.

\end{thebibliography}

\appendix
\section*{Supplementary material}
\label{sec:supp}
In Sect.~\ref{sec:synthetic} we provide additional benchmarking results on synthetic problems.
We explain the algorithms considered in
this paper in more detail in Sect.~\ref{sec:algorithms}.
Then we describe how we leverage scalable GP regression
in Sect.~\ref{sec:gps}.
We summarize the hyperparameters of \ALG in Sect.~\ref{sec:turbo_details} and give additional details
on how we shrink and expand the trust regions.
Thompson sampling is summarized in Sect.~\ref{sec:ts_details}.
Finally, we describe the test problems in Sect.~\ref{sec:test_problems} and provide runtimes for all benchmark problems in Sect.~\ref{sec:runtimes}.

\section{Synthetic experiments}
\label{sec:synthetic}
We present results on four popular synthetic problems:
Ackley with domain \smash{$[-5, 10]^{10}$}, Levy with domain \smash{$[-5, 10]^{10}$}, Rastrigin with domain
\smash{$[-3, 4]^{10}$}, and the $6$D Hartmann function with domain \smash{$[0, 1]^6$}.
The optimizers are given a budget of $50$ batches of size $q=10$ which results in a total of $n=500$ function evaluations.
All methods use $20$ initial points from a Latin hypercube design (LHD)~\citep{McKay1979} except for \ALG-$5$,
where we use $10$ initial points in each local region. To compute confidence intervals on the results, we use $30$ runs.
For \HeSBO we used target dimension~$4$ for Hartmann6 and~$6$ for the other benchmarks.

\begin{figure}[!ht]
    \centering
    \includegraphics[width=0.98\textwidth]{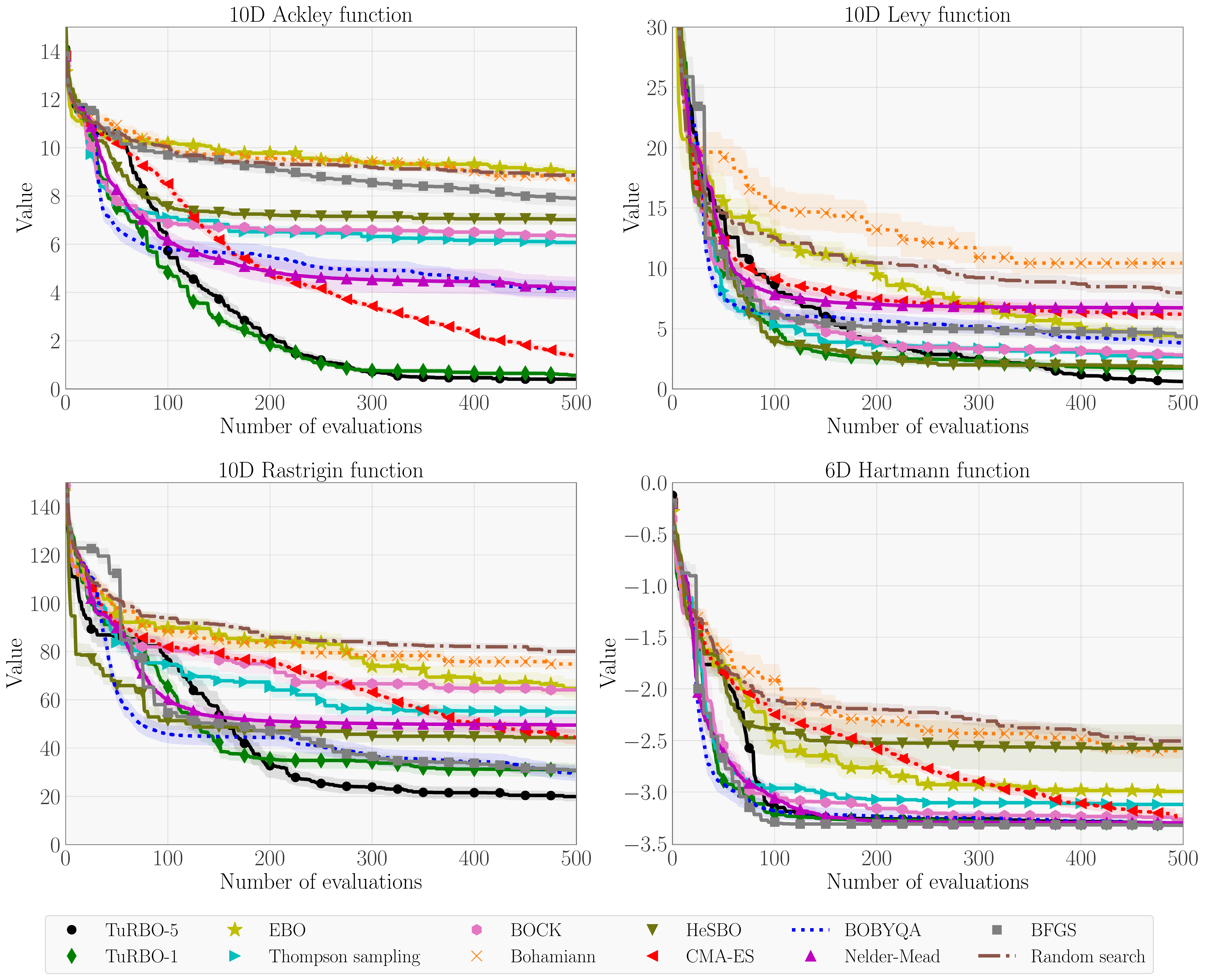}
    \caption{\ALG and \ALG-$5$ perform well on all synthetic benchmark problems.
    \HeSBO performs well on Levy and Rastrigin.
    \BOBYQA and \BFGS are competitive on Rastrigin and Hartmann6, showing that
    local optimization can outperform global optimization on multimodal functions.}
    \label{fig:synthetic}
\end{figure}

Fig.~\ref{fig:synthetic} summarizes the results. We observed a good performance for \ALG-$1$ and \ALG-$5$ on all test problems.
\ALG-$1$ and \ALG-$5$ outperform other methods on Ackley and consistently find solutions close to the global optimum.
The results for Levy also show that \ALG-$5$ clearly performs best.
However, \ALG-$1$ found solutions close to the global optimum in some trials but struggled in others,
which shows that a good starting position is important.
On Rastrigin, \ALG-$5$ performs the best. \BOBYQA and \BFGS perform comparably to \ALG-$1$.
In contrast, the $6$D Hartmann function is much easier and most methods converge quickly.

Interestingly, the embedding-based \HeSBO algorithm performs well on Levy and Rastrigin.
On the other hand, \BOHAMIANN struggles compared to other BO methods, suggesting that its model fit is inaccurate compared to GP-based methods.
We also observe that \CMA finds good solutions eventually for Ackley, Rastrigin, and Hartmann, albeit considerably slower than~\ALG.
For Levy \CMA seems stuck with suboptimal solutions.

\section{Algorithms background}
\label{sec:algorithms}
In this section, we provide additional background on the three categories of competing optimization methods: traditional local optimizers,
evolutionary algorithms, and other recent works in large-scale BO\@.
Namely, we compare \ALG to Nelder-Mead (\NM), \BOBYQA, \BFGS, \EBO, Bayesian optimization with cylindrical kernels (\BOCK), \HeSBO, \BOHAMIANN,
Thompson sampling with a global GP (\GPTS), \CMA, and random search (\RS)\@. This is an extensive set of state-of-the-art optimization
algorithms from both local and global optimization.

For local optimization, we use the popular \NM, \BOBYQA, and \BFGS methods with multiple restarts.
They are all initialized from the best of a few initial points.
We use the \texttt{Scipy}~\citep{jones2014scipy} implementations of \NM and \BFGS and the
\texttt{nlopt}~\citep{johnson2014nlopt} implementation of \BOBYQA.

Evolutionary algorithms often perform well for black-box optimization with a large number of function evaluations.
These methods are appropriate for large batch sizes since they evaluate a population in parallel.
We compare to \CMA~\citep{hansen2006cma} as it outperforms differential evolution, genetic algorithms, and particle swarms in most of our experiments.
We use the \texttt{pycma}\footnote{\url{https://github.com/CMA-ES/pycma}} implementation with the default settings and a population size equal to the batch size.
The population is initialized from the best of a few initial points.

To the best of our knowledge, \EBO is the only BO algorithm that has been applied to problems with large batch sizes and tens of thousands of evaluations.
We also compare to \GPTS, \BOCK, \HeSBO, and \BOHAMIANN, all using Thompson sampling as the acquisition function.
The original implementations of \BOCK and \BOHAMIANN often take hours to suggest a single point and do not support batch suggestions.
This necessitated changes to use them for our high-dimensional setting with large batch sizes.
To generate a discretized candidate set, we generate a set of scrambled Sobol sequences with \num{5000} points for each batch.

\section{Gaussian process regression}
\label{sec:gps}
We further provide details on both the computational scaling and modeling setup for the GP\@.
To address computational issues, we use \texttt{GPyTorch}~\citep{gardner2018gpytorch} for scalable GP regression.
\texttt{GPyTorch} follows \citet{dong2017scalable} to solve linear systems using the conjugate gradient (CG) method and approximates
the log-determinant via the Lanczos process. Without \texttt{GPyTorch}, running BO with a GP model for more than a few thousand evaluations
would be infeasible as classical approaches to GP regression scale cubically in the number of data points.

On the modeling side, the GP is parameterized using a Mat\'ern-$5/2$ kernel with ARD and a constant mean function for all experiments.
The GP hyperparameters are fitted before proposing a new batch by optimizing the log-marginal likelihood.
The domain is rescaled to \smash{$[0, 1]^d$} and the function values are standardized before fitting the GP\@.
We use a Mat\'ern-$5/2$ kernel with ARD for \ALG and use the following bounds for the hyperparameters:
(lengthscale) $\lambda_i \in [0.005, 2.0\,]$, (signal variance) $s^2 \in [0.05, 20.0]$, (noise variance) $\sigma^2 \in [0.0005, 0.1]$.

\section{\ALG details}
\label{sec:turbo_details}
In all experiments, we use the following hyperparameters for \ALG-$1$: $\tau_{\text{succ}} = 3$, $\tau_{\text{fail}} = \lceil d / q \rceil$,
$\len_{\textrm{min}} = 2^{-7}$, $\len_{\textrm{max}} = 1.6$, and $\len_{\textrm{init}} = 0.8$, where $d$ is the number of dimensions and $q$
is the batch size. Note that this assumes the domain has been scaled to the unit hypercube $[0, 1]^d$.
When using \ALG-$1$, we consider an improvement from at least one evaluation in the batch a \emph{success} \cite{regis2007stochastic}. In
this case, we increment the success counter and reset the failure counter to zero. If no point in the batch improves the current best solution
we set the success counter to zero and increment the failure counter.

When using \ALG with more than one TR, we use the same tolerances as in the sequential case ($q=1$) as the number of evaluations allocated
by each TR may differ in each batch. We use separate success and failure counters for each TR. We consider a batch a success for
$\text{TR}_{\ell}$ if $q_{\ell} > 0$ points are selected from this TR and at least one is better than the best solution in this TR.
The counters for this TR are updated just as for \ALG-$1$ in this case. If all $q_{\ell} > 0$ evaluations are worse than the current best solution we consider
this a failure and set the success counter to zero and add $q_{\ell}$ to the failure counter. The failure counter is set to $\tau_{\text{fail}}$
if we increment past this tolerance, which will trigger a halving of its side length.

For each TR, we initialize $\len \gets \len_{\text{init}}$ and terminate the TR when $\len < \len_{\min}$. Each TR in \ALG uses a
candidate set of size $\min\{100d, 5000\}$ on which we generate each Thompson sample.
We create each candidate set by first generating a scrambled Sobol sequence within the intersection of the TR and the domain $[0, 1]^d$.
A new candidate set is generated for each batch.
In order to not perturb all coordinates at once, we use the value in the Sobol sequence with probability $\min\{1, 20/d\}$ for a
given candidate and dimension, and the value of the center otherwise.
A similar strategy is used by \citet{regis2013combining} where perturbing only a few dimensions at a time showed to substantially improve the performance for high-dimensional functions.

\section{Thompson sampling}
\label{sec:ts_details}
In this section, we provide details and pseudo-code that makes the background on Thompson sampling (\TS) with GPs precise.
Conceptually, \TS~\citep{thompson1933likelihood} for BO works by drawing a function $f$ from the surrogate model (GP) posterior.
It then makes a suggestion by reporting the optimum of the function $f$.
This process is repeated independently for multiple suggestions ($q > 1$)\@.
The exploration-exploitation trade off is naturally handled by the stochasticity in sampling.

Furthermore, parallel batching is naturally handled by the marginalization coherence of GPs.
Many acquisition functions handle batching by \emph{imputing} function evaluations for the other suggested (but unobserved)
points via sampling from the posterior.
Independent \TS for parallel batches is exactly equivalent to conditioning on imputed values for unobserved suggestions.
This means \TS also trivially handles \emph{asynchronous} batch sampling~\citep{hernandez2017parallel,kandasamy2018parallelised}.

Note that we cannot sample an entire function $f$ from the GP posterior in practice.
We therefore work in a discretized setting by first drawing a finite \emph{candidate set}; this puts us in the same setting as the
traditional multi-arm bandit literature.
To do so, we sample the GP marginal on the candidate set, and then apply regular Thompson sampling.

\section{Test problems}
\label{sec:test_problems}
In this section we provide some brief additional details for the test problems.
We refer the reader to the original papers for more details.

\subsection{Robot pushing}
The robot pushing problem was first considered in \citet{wang2017batched}.
The goal is to tune a controller for two robot hands to push two objects to given target locations.
The robot controller has $d=14$ parameters that specify the location and rotation of the hands, pushing speed, moving direction, and pushing time.
The reward function is
\smash{$f(\vec x) = \sum_{i=1}^2 \|\vec x_{gi} - \vec x_{si}\| - \|\vec x_{gi} - \vec x_{fi}\|$},
where $\vec x_{si}$ are the initial positions of the objects, $\vec x_{fi}$ are the final positions of the objects, and $\vec x_{gi}$ are the goal locations.

\subsection{Rover trajectory planning}
This problem was also considered in \citet{wang2017batched}.
The goal is to optimize the trajectory of a rover over rough terrain, where the trajectory is determined by fitting a B-spline to $30$ points in a $2$D plane.
The reward function is $f(\vec x) = c(\vec x) - 10(\|\vec x_{1,2} - \vec x_s\|_1 + \|\vec x_{59,60} - \vec x_g\|_1) + 5$,
where $c(\vec x)$ penalizes any collision with an object along the trajectory by ${-20}$.
Here, $\vec x_s$ and $\vec x_g$ are the desired start and end positions of the trajectory.
The cost function hence adds a penalty when the start and end positions of the trajectory are far from the desired locations.

\subsection{Cosmological constant learning}
The cosmological constant experiment uses luminous red galaxy data from the Sloan Digital Sky Survey~\cite{tegmark2006cosmological}.
The objective function is a likelihood estimate of a simulation based astrophysics model of the observed data.
The parameters include various physical constants, such as Hubble's constant, the densities of baryonic and other
forms of matter. We use the nine parameters tuned in previous papers, plus three additional parameters chosen
from the many available to the simulator.

\subsection{Lunar lander reinforcement learning}
The lunar lander problem is taken from the OpenAI gym\footnote{\url{gym.openai.com/envs/LunarLander-v2/}}.
The objective is to learn a controller for a lunar lander that minimizes fuel consumption and distance to a landing target, while also preventing crashes.
At any time, the state of the lunar lander is its angle and position, and their respective time derivatives.
This $8$-dimensional state vector $\vec s$ is passed to a handcrafted parameterized controller that determines which of $4$ actions $a$ to take.
Each corresponds to firing a booster engine: $a \in \{\text{nothing, left, right, down}\}$.
The handcrafted control policy has $d=12$ parameters that parameterize linear score functions of the state vector and also the thresholds that determine which action to prioritize.
The objective is the average final reward over a fixed constant set of $50$ randomly generated terrains, initial positions, and initial velocities.
Simulation runs were capped at \num{1000} time steps, after which failure to land was scored as a crash.

\section{Runtimes}
\label{sec:runtimes}
In Table~\ref{tab:runtimes}, we provide the algorithmic runtime for the numerical experiments. This is the total
runtime for one optimization run, excluding the time spent evaluating the objective function.
We see that the local optimizers and the evolutionary methods run with little to no overhead on all problems.
The BO methods with a global GP model become computationally expensive when the number of evaluations increases and we leverage scalable GPs
on an NVIDIA RTX $2080$ TI.
\ALG does not only outperform the other BO methods, but runs in minutes on all test problems and is in fact more than $2000\times$
faster than the slowest BO method.

\begin{table}[!ht]
    \centering
    \resizebox{0.98\textwidth}{!}{
        \begin{tabular}{|l|r|r|r|r|r|r|}
            \hline
            & Synthetic & Lunar landing & Cosmological constant & Robot pushing & Rover trajectory & Ackley-$200$ \\
            \hline
            Evaluations $n$ & \num{500} & \num{1500} & \num{2000} & \num{10000} & \num{20000} & \num{10000} \\
            Dimensions $d$ & \num{6} or \num{10} & \num{12}  & \num{12} & \num{14} & \num{60} & \num{200} \\
            \hline
            \ALG       & \SI{< 1}{\minute} & \SI{< 1}{\minute}  & \SI{< 1}{\minute} & \SI{8}{\minute}   & \SI{22}{\minute}  & \SI{10}{\minute}  \\
            \EBO       & \SI{4}{\minute}   & \SI{23}{\minute}   & \SI{1}{\hour}     & \SI{11}{\day}     & \SI{> 30}{\day}   & NA                \\
            \GPTS      & \SI{3}{\minute}   & \SI{6}{\minute}    & \SI{11}{\minute}  & \SI{1}{\hour}     & \SI{3}{\hour}     & \SI{1}{\hour}     \\
            \BOCK      & \SI{6}{\minute}   & \SI{10}{\minute}   & \SI{19}{\minute}  & \SI{2}{\hour}     & \SI{7}{\hour}     & \SI{2}{\hour}     \\
            \BOHAMIANN & \SI{2}{\hour}     & \SI{5}{\hour}      & \SI{7}{\hour}     & \SI{20}{\hour}    & \SI{2}{\day}      & \SI{25}{\hour}    \\
            \NM        & \SI{< 1}{\minute} & \SI{< 1}{\minute}  & \SI{< 1}{\minute} & \SI{< 1}{\minute} & \SI{< 1}{\minute} & \SI{< 1}{\minute} \\
            \CMA       & \SI{< 1}{\minute} & \SI{< 1}{\minute}  & \SI{< 1}{\minute} & \SI{< 1}{\minute} & \SI{< 1}{\minute} & \SI{< 1}{\minute} \\
            \BOBYQA    & \SI{< 1}{\minute} & \SI{< 1}{\minute}  & \SI{< 1}{\minute} & \SI{< 1}{\minute} & \SI{< 1}{\minute} & \SI{< 1}{\minute} \\
            \BFGS      & \SI{< 1}{\minute} & \SI{< 1}{\minute}  & \SI{< 1}{\minute} & \SI{< 1}{\minute} & \SI{< 1}{\minute} & \SI{< 1}{\minute} \\
            \RS        & \SI{< 1}{\minute} & \SI{< 1}{\minute}  & \SI{< 1}{\minute} & \SI{< 1}{\minute} & \SI{< 1}{\minute} & \SI{< 1}{\minute} \\
            \hline
        \end{tabular}
    }
    \vspace{3mm}
    \caption{Algorithmic overhead for one optimization run for each test problem.
    The times are rounded to minutes, hours, or days.}
    \label{tab:runtimes}
\end{table}

\end{document}